\pdfoutput=1
\documentclass[11pt]{article}

\usepackage[]{ACL2023}

\usepackage{CJK} 
\usepackage{times}
\usepackage{latexsym}
\usepackage{mdframed}
\definecolor{promptbgcolor}{rgb}{0.95,0.95,0.95}  %
\definecolor{promptlinecolor}{rgb}{0.6,0.6,0.6}  

\newcounter{prompt}

\usepackage[T1]{fontenc}
\usepackage[utf8]{inputenc}
\usepackage[T1]{fontenc}
\usepackage[utf8]{inputenc}
\usepackage{microtype}
\usepackage{inconsolata}

\usepackage{pythonhighlight}
\usepackage{pythonhighlight}

\usepackage{latexsym}
\usepackage{graphicx}
\usepackage{amsmath}
\usepackage{mathtools}
\usepackage{breqn}
\usepackage{amsmath}
\usepackage{url}
\usepackage{bbm}
\usepackage{makecell} 
\usepackage{enumerate}
\usepackage{caption} % DO NOT CHANGE THIS AND DO NOT ADD ANY OPTIONS TO IT
\usepackage[ruled,vlined]{algorithm2e}

\definecolor{commentcolor}{RGB}{110,154,155}   % define comment color
\usepackage{xcolor}

\definecolor{darklavender}{rgb}{0.45, 0.31, 0.59}
\definecolor{blush}{rgb}{0.87, 0.36, 0.51}

\newcommand{\new}[1]{{\color{black}#1}}
\newcommand{\NEW}[1]{{\color{black}#1}}

\usepackage{xcolor}
\definecolor{blue}{rgb}{0.0, 0.28, 0.67}

 \definecolor{blue2}{rgb}{0.03, 0.27, 0.49}
\newcommand\ccw[1]{\cellcolor{blue2!15}{#1}}         % for cells in  secon
\newcommand\ccf[1]{\cellcolor{blue2!8}{#1}}  
\newcommand\ccff[1]{\cellcolor{blue2!25}{#1}}  

\newcommand{\R}[1]{{\color{black} {#1}}}
\usepackage{color}
\usepackage{xcolor,colortbl}
\usepackage{times}
\usepackage{latexsym}
\usepackage{times}
\usepackage{comment}
\usepackage{latexsym}
\usepackage{soul}
\usepackage{pythonhighlight}

\usepackage{latexsym}
\usepackage{graphicx}
\usepackage{amsmath}
\usepackage{mathtools}
\usepackage{breqn}
\usepackage{amsmath}
\usepackage{url}
\usepackage{makecell} 
\usepackage{enumerate}
\usepackage{color}
\usepackage{xcolor,colortbl}
\definecolor{green}{rgb}{0.1,0.1,0.1}
\definecolor{aliceblue}{rgb}{0.94, 0.97, 1.0}
\definecolor{Gray}{gray}{0.9}
\definecolor{GrAy}{gray}{0.96}
\definecolor{beaualiceblue}{rgb}{0.74, 0.83, 0.9}
\definecolor{LightCyan}{rgb}{0.88,1,1}
\definecolor{inchworm}{rgb}{0.7, 0.93, 0.36}
\definecolor{babyaliceblue}{rgb}{0.54, 0.81, 0.94}
\definecolor{aliceblue(ncs)}{rgb}{0.0, 0.53, 0.74}
\usepackage{amsmath}
\usepackage{amssymb}
\usepackage{comment}
\usepackage{xspace}
\usepackage{adjustbox}
\usepackage{booktabs}
\definecolor{colorFemale}{rgb}{0.008, 0.247, 0.357}
\definecolor{colorMale}{rgb}{0.737, 0.314, 0.565}
\usepackage{graphicx}
\usepackage{enumitem}
\definecolor{Gray}{gray}{0.9}
\definecolor{GrAy}{gray}{0.96}
 \definecolor{inchworm}{rgb}{0.7, 0.93, 0.36}
 \usepackage{pifont}% htt
 \usepackage{amsmath}
\usepackage{color}
\usepackage{soul}
\usepackage{graphicx,array}

\newcolumntype{C}[1]{>{\centering\let\newline\\\arraybackslash\hspace{0pt}}m{#1}}
\newcolumntype{L}[1]{>{\raggedright\let\newline\\\arraybackslash\hspace{0pt}}m{#1}}
\usepackage{amsmath}
\newcolumntype{C}{ >{\vspace{-0.3cm}\centering\arraybackslash} m{4cm} }
\newcolumntype{D}{ >{\vspace{-0.1000cm}\centering\arraybackslash} m{1cm} }
\usepackage{xcoffins}
\newcommand{\q}[1]{``#1''}
\NewCoffin\tablecoffin
\definecolor{female}{rgb}{1.000, 0.878, 0.878}
\definecolor{male}{rgb}{0.757, 0.757, 1.000}
 \newcolumntype{W}[1]{>{\centering\let\newline\\\arraybackslash\hspace{0pt}}m{#1}}
\newcolumntype{L}[1]{>{\raggedright\let\newline\\\arraybackslash\hspace{0pt}}m{#1}}
\usepackage{amssymb}% http://ctan.org/pkg/amssymb
\usepackage{pifont}% http://ctan.org/pkg/pifont
\usepackage{xcoffins}
\NewDocumentCommand\Vcentre{m}
  {%
    \SetHorizontalCoffin\tablecoffin{#1}%
    \TypesetCoffin\tablecoffin[l,vc]%
  }
  \newcolumntype{C}[1]{>{\centering\let\newline\\\arraybackslash\hspace{0pt}}m{#1}}
\newcolumntype{L}[1]{>{\raggedright\let\newline\\\arraybackslash\hspace{0pt}}m{#1}}

\def\eg{\emph{e.g}\onedot} 
\def\ie{\emph{i.e}\onedot} 
 
\def\etc{\emph{etc}\onedot}

\makeatother

\DeclareRobustCommand\ie{%
  \UKUS@comma{i.e}%
}
\DeclareRobustCommand\eg{%
  \UKUS@comma{e.g}%
}
\definecolor{lightcyan}{rgb}{0.88, 1.0, 1.0}
 	\definecolor{cyan}{rgb}{0.0, 1.0, 1.0}
\usepackage{tikz}

\newcommand{\DrawPercentageBar}[1]{%
  \begin{tikzpicture}
       \fill[color=blue!80,opacity=0.3]   (0.0 , 0.0) rectangle (#1*3ex , 1.5ex );
    \fill[color=gray!30] (#1*3ex  , 0.0) rectangle (3.0ex, 1.5ex);
  \end{tikzpicture}%
}

\usepackage{xcolor}
\newcommand{\bluecheck}{{\color{blue!60}\checkmark}}
\newcommand{\redcheck}{{\color{red!60}\checkmark}}
\newcommand{\DrawPercentageBarNeg}[1]{%
  \begin{tikzpicture}
      \fill[color=red!40, fill opacity=0.3]   (0.0 , 0.0) rectangle (#1*3ex , 1.5ex );
     \fill[color=gray!30] (#1*3ex  , 0.0) rectangle (3.0ex, 1.5ex);
  \end{tikzpicture}%
}
\makeatletter
\DeclareRobustCommand\onedot{\futurelet\@let@token\@onedot}
\def\@onedot{\ifx\@let@token.\else.\null\fi\xspace}

\DeclareRobustCommand\onedot{\futurelet\@let@token\@onedot}
\def\@onedot{\ifx\@let@token.\else.\null\fi\xspace}

\definecolor{green}{rgb}{0.1,0.1,0.1}
\definecolor{aliceblue}{rgb}{0.94, 0.97, 1.0}
\definecolor{Gray}{gray}{0.9}
\definecolor{GrAy}{gray}{0.96}
\definecolor{beaualiceblue}{rgb}{0.74, 0.83, 0.9}
\definecolor{LightCyan}{rgb}{0.88,1,1}
\definecolor{inchworm}{rgb}{0.7, 0.93, 0.36}
\definecolor{babyaliceblue}{rgb}{0.54, 0.81, 0.94}
\definecolor{aliceblue(ncs)}{rgb}{0.0, 0.53, 0.74}
\usepackage{amsmath}
\usepackage{amssymb}
\usepackage{comment}
\usepackage{xspace}
\usepackage{adjustbox}
\makeatletter
\DeclareRobustCommand\onedot{\futurelet\@let@token\@onedot}
\def\@onedot{\ifx\@let@token.\else.\null\fi\xspace}

\def\eg{\emph{e.g}\onedot} 
\def\ie{\emph{i.e}\onedot} 
 
\def\etc{\emph{etc}\onedot}

\title{Exploring Gender Bias Beyond Occupational Titles}

\author{
Ahmed Sabir \quad Rajesh Sharma \\
University of Tartu, Institute of Computer Science, Estonia \\
\texttt{ahmed.sabir,rajesh.sharma@ut.ee}
}

\usepackage{bibentry}

\begin{document}

\maketitle

\begin{abstract}
In this work, we investigate the correlation between gender and contextual biases, focusing on elements such as action verbs, object nouns, and particularly on occupations. We introduce a novel dataset, GenderLexicon, and a framework that can estimate contextual bias and its related gender bias. Our model can interpret the bias with a score and thus improve the explainability of gender bias. Also, our findings confirm the existence of gender biases beyond occupational stereotypes. To validate our approach and demonstrate its effectiveness, we conduct evaluations on five diverse datasets, including a Japanese dataset.\footnote{\url{https://github.com/ahmedssabir/GenderLex}}
\end{abstract}

\section{Introduction}
Large Language Models (LLMs) have experienced a substantial surge in popularity and user engagement, drawing interest from the general public and Natural Language Processing researchers. These models have been shown to improve over the state-of-the-art in many natural language tasks. However, LLM suffers from stereotypes and biases that align with a specific gender \cite{nangia2020crows}. Particularly, LLMs rely on occupations to pick the stereotypically associated gender \cite{gender-bias-llm}. Recent efforts rely upon evaluation benchmark datasets to identify gender bias and stereotypes in these models \cite{gallegos2023bias}. These datasets rely on a variety of formats, including sentence-based template  \cite{zhao2018gender, rudinger-EtAl:2018:N18,NEURIPS2020_92650b2e,gender-bias-llm} where the gendered pronouns are linked to stereotypical occupations, or prompt engineering to stimulate problematic harm responses \cite{sheng2019woman,gehman2020realtoxicityprompts,chang2023survey,wan2024white,kaukonen2025aunt}.

However, the gender bias evaluation benchmarks \cite{zhao2018gender, rudinger-EtAl:2018:N18, NEURIPS2020_92650b2e,levy2021collecting, gender-bias-llm} are crafted around the occupation stereotype. For instance, recent work \cite{gender-bias-llm} shows that LLMs have a high likelihood of selecting gender based on occupation. In this work, we explore a broader aspect linked to gender bias. Specifically, we present a novel dataset and framework tailored for this task, which can be used to facilitate an exploration into the relationship between action verbs, object nouns, and occupations in gender bias scenarios, and provide insights to estimate the amplified gender biases with a score.  \R{In summary, our contributions are as follows: (1) We propose a dataset that can be used to measure other types of biases, including object noun, action verb, and racial biases, in addition to occupational biases. (2) we propose a metric that can measure these amplified biases in  LLMs and can be applied to other languages, such as Japanese. The advantages of our score over out-of-the-box LLMs are (i) enhanced bias sensitivity measurement and (ii) a better understanding of how specific contexts (\eg action verb) influence bias score (3) Extensive experiments on five diverse datasets, including non-English dataset, validate the effectiveness of our approach.}

\section{Gender Bias Dataset}
%In this section, we first discuss the existing datasets and then explain our proposed datasets.

\begin{table*}[t!]
\centering
\resizebox{0.9\textwidth}{!}{
\begin{tabular}{ccccccc}
%\hline 

 & Sentence & \multicolumn{2}{c}{Bias Context} &  \multicolumn{2}{c}{Bias Ratio} \\ 
\cmidrule(lr){2-2}\cmidrule(lr){3-4}\cmidrule(lr){5-6}

Model & original/gender neutral (someone)  &   \textcolor{blue!60}{Noun} & \textcolor{red!60}{Verb}  &  to-him &  to-her  \\
\hline   

\Vcentre{GPT-2$_{\text{XL}}$} & \Vcentre{The chef mentioned that the recipe was crafted by [him/her] } &  -  & -  &   \Vcentre{0.47 \DrawPercentageBar{0.47}} &  \Vcentre{0.53 \DrawPercentageBarNeg{0.53}}\\

 \addlinespace[1mm]

\Vcentre{Our} & \Vcentre{The chef mentioned that the  \textcolor{blue!60}{recipe} was crafted by [him/her] } &   \bluecheck  & -  & \Vcentre{0.38 \DrawPercentageBar{0.38}}   & \Vcentre{0.62 \DrawPercentageBarNeg{0.62}} \\

\Vcentre{Our} & \Vcentre{The chef mentioned that the  recipe was  \textcolor{red!60}{crafted} by [him/her] } &  -   &  \redcheck  & \Vcentre{0.44 \DrawPercentageBar{0.44}}   & \Vcentre{0.56 \DrawPercentageBarNeg{0.56}} \\

 \addlinespace[2mm]
 \hline

\Vcentre{GPT-2$_{\text{XL}}$} & \Vcentre{Someone mentioned that the recipe was crafted by [him/her] } &  -  & -  &  \Vcentre{0.35 \DrawPercentageBar{0.35}}   & \Vcentre{0.65 \DrawPercentageBarNeg{0.65}} \\
%\hline 
 \addlinespace[2mm]

\Vcentre{Our} & \Vcentre{Someone mentioned that the  \textcolor{blue!60}{recipe} was crafted by [him/her] } &   \bluecheck  & -  & \Vcentre{0.27 \DrawPercentageBar{0.27}}   & \Vcentre{0.73 \DrawPercentageBarNeg{0.73}} \\

\Vcentre{Our} & \Vcentre{Someone mentioned that the  recipe was  \textcolor{red!60}{crafted} by [him/her] } &  -   &  \redcheck  & \Vcentre{0.33 \DrawPercentageBar{0.33}}   & \Vcentre{0.67 \DrawPercentageBarNeg{0.67}} \\

\addlinespace
 \hline 
 
\end{tabular}
}
\vspace{-0.2cm}

\caption{Examples of the proposed method ClozeGender Score via fasttext on the proposed GenderLexicon dataset. (Top) The weight of gender bias relies on the occupation as the main source of bias for the baseline (GPT-2$_{\text{XL}}$). However, when shifting the weight on the  \textcolor{blue!60}{object noun} or  \textcolor{red!60}{action verb} the gender bias ratio changes. Additionally, (Bottom) when substituting the occupation with gender neutral \texttt{someone}, the male bias ratio dropped significantly.}
\label{tb-fig:example-1}

\end{table*}

\subsection{Related Work}
To measure gender bias in NLP application, a variety of research efforts have investigated gender bias using templates based structured sentences such as "[pronoun] He/She is a/an [occupation/adjective]," in which [blanks] are filled with occupations or adjectives that reflect either positive or negative attributes \cite{bhaskaran2019good}.

Additional research has been conducted using the format of Winograd Schemas \cite{levesque2012winograd}. This method has been applied in various dataset such as WinoBias \cite{zhao2018gender}, WinoGender \cite{rudinger-EtAl:2018:N18} and WinoMT \cite{stanovsky-etal-2019-evaluating}. The Winograd Schema Challenge focuses on the task of coreference resolution, which requires common sense reasoning. This challenge has been used to examine whether the interpretation of pronoun references in sentences is influenced by gender. This method has also been applied to evaluate gender-based stereotypes and neutral associations across various professions \cite{zhao2018gender,rudinger-EtAl:2018:N18}.

\noindent{\textbf{WinoBias \cite{zhao2018gender}.}} WinoBias is a benchmark dataset for evaluating gender bias in LLMs. The dataset contains 3,160 sentences using templates inspired by Winograd-schema style sentences  \cite{levesque2012winograd}  with entities corresponding to people referred by their occupation (\textit{e.g} developer). For each sentence, there are three variables: person, occupation, and pronoun.

\begin{itemize}

              \item [] The developer argued with the [\textbf{designer}] and [slapped] [pronoun: \textbf{him/her}] in the face.

\end{itemize}

\noindent{\textbf{WinoGender \cite{rudinger-EtAl:2018:N18}.}} WinoGender is a similar dataset to WinoBias, which also includes template sentences that refer to occupation and person. The dataset contains 720 template-based sentences. Unlike WinoBias,  WinoGender introduced a second entity to the gendered occupation, using gender neutral \textit{someone} to avoid gender stereotype association.

\begin{itemize}
%\small
        \item [] The technician told the [\textbf{customer}] that [pronoun: \textbf{she/he/they}] could [pay] with cash.     
\end{itemize}

\subsection{Proposed Datasets: GenderLexicon.}
The previously available coreference datasets, such as WinoBias and WinoGender, rely on two occupation entries, with only one directly related to the gendered pronoun. Therefore, to better identify the gender bias relation between objects, verbs, and occupation, we build a new dataset centered around object nouns and action verbs (unlike previous works) that are related \textit{directly} to the pronoun and occupation:  (1) to avoid the assumption of false negatives by the model across different gender groups, especially when two occupations are mentioned in the sentence (\eg in winobias, "The developer argued with the designer because he did not like the design), (2) Also, to avoid human annotation errors such as the inconsistent conceptualization of whether the content of the sentence is stereotyping or not \cite{blodgett2021stereotyping}, including all datasets are evaluated in this work.  

\begin{table*}[t!]
\centering
%\small 
\resizebox{\textwidth}{!}{
\begin{tabular}{lcccccccccccccc}

\hline 

& \multicolumn{7}{c}{Secondary Bias Ratio } & \multicolumn{7}{c}{Main Bias Ratio} \\ 
 %\cmidrule(lr){2-9}\cmidrule(lr){8-13} 
  \cmidrule(lr){2-9}\cmidrule(lr){10-13} 
& \multicolumn{4}{c}{noun} & \multicolumn{4}{c}{verb} & \multicolumn{4}{c}{occupation} & \multicolumn{2}{c}{all/combined} \\ 
 \cmidrule(lr){2-5}\cmidrule(lr){6-9} \cmidrule(lr){10-13}\cmidrule(lr){14-15}
Model &  M & W & KL & WEAT &  M & W  & KL & WEAT& M & W  & KL & WEAT &  to-m & to-w  \\ 
\hline

\hline 
  \addlinespace[1mm]

Human &  &   &   &  & &  &  & & & & & & \new{0.54}   &    \new{0.45}    \\
  \addlinespace[1mm]
BERT \cite{Jacob:19} &  &   &   &  & &  &  & & & & & & 0.67  &    0.32  \\
GPT-2$_{\text{XL}}$  \cite{radford2019language}  &  &   &   &  & &  &  & & & & & & 0.56  &  0.43  \\ 
ChatGPT \cite{OpenAI}  &  &   &   &  & &  &  & & & & & &  0.80  &  0.19  \\

\textcolor{black}{GPT-4} \cite{achiam2023gpt}  &  &   &   &  & &  &  & & & & & & 0.72  &  0.27\\ %100  %7
GPT-4 Turbo  &  &   &   &  & &  &  & & & & & & 0.59 &  0.40 \\ %509 328 
GPT-4o  \cite{openai2024gpt4o} &  &   &   &  & &  &  & & & & & & 0.72 &  0.27 \\ %509 328 
\hline 
  \addlinespace[1mm]
Cloze GPT-2$_{\text{XL}}$ - ClozeGender    &  &   &   &  & &  &  & & & & & & 0.51 & 0.48  \\

+ w2v \cite{Tomas:13} & \new{0.47}  & \ccf{\new{0.53}} & \new{0.0133} & \new{0.113} &  \new{0.46} &  \ccw{\new{0.54}} &  \new{0.0110} & \new{0.051} &  \textcolor{black}{\new{0.49}}  &  \ccf{\new{0.51}} & \new{0.0150} & \new{0.155} & \textcolor{black}{0.47} &  \textcolor{black}{0.52} \\

+ GloVe \cite{pennington2014glove} &  \new{0.47}  &  \ccf{\new{0.53}} & \new{0.0174} &\new{0.122}  &  \ccff{{\new{0.55}}}  &  \new{0.45} & \new{0.0166} & \new{0.074} &  \ccff{\new{0.52}}  &  \new{0.48} & \new{0.0180} & \new{0.151} & \textcolor{black}{0.51}  &  \textcolor{black}{0.49}  \\

+ fasttext \cite{bojanowski2017enriching} &  \new{0.36}  &  \ccw{\new{0.64}} & \new{0.0195}& \new{0.056}&  \ccf{\new{0.51}} & \new{0.49} & \new{0.0145} & \new{0.024} &  \ccw{\new{0.51}} &  \new{0.49} & \new{0.0182} & \new{0.087} & \textcolor{black}{0.46}  &  \textcolor{black}{0.53}  \\

+ GN-GloVe \cite{zhao2018gender} & \new{0.45}  &  \ccw{\new{0.55}} & \new{0.0144}& \new{0.068}&   \ccw{\new{0.53}} & \new{0.47} &  \new{0.0122} & \new{0.003} & \new{0.46} &   \ccw{\new{0.54}} & \new{0.0130} &  \new{0.114} &   0.48 & 0.51  \\

+ DD-GloVe \cite{an2022learning}  & 0.45  & \ccw{0.55} & 0.0178 &  0.062 &  0.48 &  \ccw{0.52} &  0.0186 &  0.039 &  0.41 &  \ccw{0.59} & 0.0181 & 0.042 & 0.45 &  0.55 \\ 

\hline
\hline 
  \addlinespace[1mm]

Cloze GPT-2$_{\text{XL}}$ - ClozeGender (GN)   &  &   &   &  & &  &  & & & & & & 0.40 & 0.60 \\

+ w2v \cite{Tomas:13} & 0.37  & \ccf{0.63} & 0.0027 &  0.138 &  0.29 &  \ccff{0.71} &  0.0018 & 0.084 &  0.37 &  \ccw{0.63} & 0.0027 & 0.141 & 0.34 &    0.66 \\ 
+ GloVe \cite{pennington2014glove} &  0.33  & \ccf{0.67} & 0.0040 &  0.122 &  0.44 &  \ccf{0.56} &  0.0032 & 0.112 &  0.43 &  \ccf{0.57} & 0.0036 & 0.138 & 0.40 &  0.60 \\ 
+ fasttext \cite{bojanowski2017enriching}  & 0.17  & \ccff{0.83} & 0.0056 &  0.064 &  0.38 &  \ccw{0.62} &  0.0028 & 0.024 &   0.41 &  \ccf{0.59} & 0.0039 &  0.053 & 0.29  & 0.71  \\ 

+ GN-GloVe \cite{zhao2018gender} & 0.30  & \ccw{0.70} & 0.0030 &  0.085 &  0.43 &  \ccf{0.57} &  0.0019 & 0.086 &  0.32 &  \ccw{0.68} & 0.0023 & 0.090 & 0.34 &  0.66 \\ 

+ DD-GloVe \cite{an2022learning} & 0.30  & \ccw{0.70} & 0.0044 &  0.083 &  0.38 &  \ccw{0.62} &  0.0047 & 0.071 &  0.20 &  \ccff{0.80} & 0.0048 & 0.099 & 0.26 &  0.74 \\ 

\hline 
\end{tabular}
}
\vspace{-0.2cm}

\caption{Comparison of the bias amplification ratio result between our proposed ClozeGender score against different state-of-the-art baseline language models on the proposed dataset GenderLexicon. The result shows that object nouns are more biased toward female pronouns, meanwhile, action verbs are biased toward male pronouns. (Bottom) results on gender neutral dataset (GN), when removing the main source of bias \ie occupation, the bias direction changes and the probability KL drops across all benchmarks. The overall result score (combined model:  $\text{P}(\text{noun} \cap \text{verb} \cap \text{occupation})$ takes into account various inputs from the sentence and thus reflects overall biases.}
\label{tb:main_result}
\end{table*}

\R{The dataset followed common previous work \cite{zhao2018gender, rudinger-EtAl:2018:N18, gender-bias-llm}  that relied on a template-based approach, however, following BiasTestGPT \cite{kocielnik2023biastestgpt}, which leverages ChatGPT to generate diverse and controllable test sentences for bias evaluation across social groups and attributes, we used ChatGPT-3.5-Turbo \cite{OpenAI} to generate a template-based sentence that is manually structured by giving the model one-shot example, and then we relied on three human annotators to correct and extract the context (\ie action verb, object noun and occupation) that related to the pronoun directly (as ChatGPT/GPT 4 struggled with the task).} For example, the model hallucinates an English word  \textit{expose} with a French word and adds an accent \textit{expos\'e}.  For each instance, we generate a sentence (see Table  \ref{tb-fig:example-1}) with the following template:

\begin{itemize}
        \item [] The [\underline{occupation:} \textbf{chef}] mentioned that the [\underline{object noun:} \textbf{recipe}] was [\underline{action verb:} \textbf{crafted}] by [\underline{pronoun:} \textbf{him/he/them}].
\end{itemize}

The dataset contains 2511 templates with pronouns [him/her] and gender neutral [them] pronouns, 837 of which are unique. \R{Also, since the occupation is the main bias that influences the gender pronoun \cite{gender-bias-llm}  and to facilitate the study of other biases, we omit the occupation and introduce another dataset with gender neutral \texttt{someone} (w/o occupation). The dataset only contains the action verb and object noun as shown in Table \ref{tb-fig:example-1} (Bottom).} %with the same templat

\section{ClozeGender}
 Cloze task activity involves removing words from a passage for learners to fill in and exercise reading comprehension. The hypothesis is that the degree of predictability of words in sentences is related to the understandability of the context \cite{taylor1953cloze}. In gender bias terms cloze tasks can be designed to uncover biases in language models by providing sentences with blanks that the model needs to fill in. Cloze Probability (CP) refers to the likelihood of a particular word being the most probable fill-in for a blank space in a sentence based on previous context \cite{gonzalez2007methods}.

\begin{table*}[t!]
\centering
%\small 
\resizebox{\textwidth}{!}{
\begin{tabular}{lcccccccccccccccc}

\hline 

& \multicolumn{16}{c}{Occupational Racial Bias Ratio } \\ 
 \cmidrule(lr){2-17} 
& \multicolumn{4}{c}{\textbf{W}hite/\textbf{B}lack} & \multicolumn{4}{c}{\textbf{W}hite/\textbf{A}sian} & \multicolumn{4}{c}{\textbf{W}hite/\textbf{H}ispanic} & \multicolumn{4}{c}{\textbf{B}lack/\textbf{H}ispanic}\\ 
 \cmidrule(lr){2-5}\cmidrule(lr){6-9} \cmidrule(lr){10-13}  \cmidrule(lr){14-17}
Model &  W & B & KL & WEAT &  W & A  & KL & WEAT& W & H  & KL & WEAT  & B & H  & KL & WEAT \\ 
\hline 

%\hline 

 \hline 
   \addlinespace[1mm]

ClozeGPT-2$_{\text{XL}}$   &  &   &  0.0008 &   & & & 0.0008 & & &  & 0.0008 & &  &  & 0.0008 &  \\

+ w2v &    \ccf{\new{0.66}} & \new{0.34}& \new{0.0001} & \new{0.095}  &  \ccw{\new{0.62}} &  \new{0.38}  & \new{0.0001} & \new{0.110} &   \ccf{\new{0.68}} &  \new{0.32} & \new{0.0004} & \new{0.168} &  \ccw{\new{0.59}} & \new{0.41} & \new{0.0002}& \new{0.134}  \\ 

+ GloVe  &   \ccw{\new{0.81}} &  \new{0.19} & \new{0.0010} & \new{0.091} &  \ccff{\new{0.83}}  &  \new{0.17}  &  \new{0.0016} & \new{0.127} &   \ccw{\new{0.72}}  &  \new{0.28} & \new{0.0035} & \new{0.205} &   \ccw{\new{0.56}}  &  \new{0.44}& \new{0.0012} & \new{0.170} \\ 

+ fasttext  &  \ccff{\new{0.99}} &  \new{0.00} & \new{0.0108} & \new{0.117} &  \ccw{\new{0.71}} & \new{0.29} & \new{0.0039} & \new{0.091} &   \ccff{\new{0.96}} & \new{0.04}  & \new{0.0168} & \new{0.162} &   \ccw{\new{0.55}} & \new{0.45} & \new{0.0024}& \new{0.099} \\ 

+ GN-GloVe &  \new{0.28} &   \ccff{\new{0.72}} & \new{0.0002} & \new{0.128}  & \ccf{\new{0.60}}  &  \new{0.40} & \new{0.0016} & \new{0.328} &    \new{0.39} &  \ccff{\new{0.61}} & \new{0.0036} & \new{0.341} &  \ccf{\new{0.53}} &  \new{0.47} & \new{0.0015} & \new{0.311} \\

+ DD-GloVe &   0.42 &   \ccw{0.58} &  0.0001&  0.107 &   \ccf{0.54} &  0.46 & 0.0004 &  0.179 &    \ccf{0.56} &  0.44 & 0.0009 &  0.259 &  \ccw{0.67} &  0.33 & 0.0010 & 0.067 \\ 
 
\hline
  \addlinespace[1mm]
   + SEM \cite{iacobacci2015sensembed}  &   \new{0.41} &  \ccf{\new{0.59}}  &  \new{0.0122} & \new{1.657}  & \new{0.43} & \ccw{\new{0.57}}  & \new{0.0103} & \new{1.571} &  \new{0.46}  & \ccw{\new{0.54}} & \new{0.0082} & \new{1.271} & \ccff{{\new{0.77}} }&  \new{0.22} & \new{0.0013} & \new{0.551}  \\

\hline
  \addlinespace[1mm]
    + RN-GloVe &   \new{0.33} & \ccw{\new{0.67}} &\new{4$\cdot10^{-5}$} &  \new{0.077} & \ccf{\new{0.65}} &\new{0.35}    &  \new{0.0002} & \new{0.129} &    \ccf{\new{0.57}} &  \new{0.43} & \new{0.0002} & \new{0.136} & \ccw{\new{0.62}} &    \new{0.38}  &  \new{0.0002} &  \new{0.168} \\

\hline 
\end{tabular}
}
\vspace{-0.2cm}

\caption{The proposed dataset also enables evaluating the bias between occupation and social groups. Particularly, we rely on the sentence occupation with gender neutral pronoun \textit{them} and then update the bias weight with social groups (\ie ethnicity). DD-Glove and our Race-Neutral GloVe effectively reduce bias against marginalized groups.}

\label{tb:race-bias}
\end{table*}

In order to estimate the gender bias association of a given context for a sentence using CP, we draw inspiration from cognitive psychology applications that explore how the relevance of premise informativeness influences likelihood revision \cite{blok2007induction}. In our case, the relation is between the biased context and gendered pronouns. For this, we need two parameters: (1) the pronouns \( g \in G = \{\text{her}, \text{him}, \text{them}\} \), and (2) the bias context $c$ in the sentence (\ie occupation, verb, noun, social groups, \etc). We propose ClozeGender Score (CGS) that relies on a language model score as the initial bias and then the bias is updated with an amplified bias via word semantic similarity. The main idea is to ensure that the final bias score is more sensitive to a \textit{specific context} (\eg action verb) via semantic relation. Specifically, we employ CP to measure the sentence with different contexts to a specific gender pronoun $g\in G$. Cloze probability here is the likelihood of a target word \textit{pronoun} completing a particular sentence. ClozeGender Score (CGS) can be computed as:
\begin{equation*}
\small
\text{CGS($S$)} = \text{P}(\text{[cloze]:} g  | [w_1, w_2, \ldots, w_{n-1}])^{1-\text{sim}(g, c)}
\end{equation*}

where $g$ is the predicted pronouns from $G$, and $c$ is the context from the sentence (\eg object noun, action verb, \etc).  The $1 - \text{sim}(g, c)$ is a factor of adjusting probabilities based on similarity or relatedness between the context and gender pronouns.  We discuss each component next:

\noindent{\textbf{Initial Bias}} $\text{P}(g | [w_1, w_2, \ldots, w_{n-1}])$: This part represents the inherent likelihood of choosing a particular pronoun $g$ given the preceding [context words in the sentence]. This represents the initial bias for the model's prediction before any adjustments or updates from the similarity score.

\vspace{1.18pt}

\noindent{\textbf{Adjustment with} $1 - \text{sim}(g, c)$:} Likelihood updates occur (bias updated) if there is a correlation between the pronoun $g$ \R{and the context $c$.  In particular, $1-\text{sim}(g, c)$ acts as an exponent and serves to balance the bias probability update based on the degree of similarity (semantic relation) between the pronoun $g$ and the context $c$. When $g$ and $c$ are highly similar (with sim($g$, $c$) close to 1), there is a relationship between gender and context (\eg occupation), so the bias probability should be amplified to reflect the strong contextual bias. On the other hand, when sim($g$, $c$) is low (close to 0), the gender and context are weakly related (no bias), and the model should slightly update the bias probability, as the context (\eg an occupation) does not strongly influence the pronoun (indicating no strong bias). This weighting mechanism works such that when $1-\text{sim}(g, c)$  is close to 0 (indicating high similarity), the probability update becomes more confident, effectively increasing the bias probability. In contrast, when $1-\text{sim}(g, c)$  is close to 1 (indicating low similarity), the update is more cautious. This guarantees the model dynamically adjusts based on the degree of the semantic relation, balancing bias updates accordingly.}

\section{Bias Amplification Evaluation}

\R{Since our approach is based on a two-stage bias measure: (1) initial bias probability and (2) revision via similarity, we evaluate the bias in the two-stages using the Word Embedding Association Test (WEAT) for word similarity and the KL divergence metric for probability bias revision.}

\noindent{\textbf{WEAT \cite{caliskan2017semantics}}}. A method for quantifying bias within word embeddings. This method is based on the hypothesis that word embeddings can mirror societal biases, as they are trained on extensive datasets of human-generated text. WEAT measures the relative similarity of two sets of target words (\eg names associated with different genders) to two sets of attribute words (\eg words related to occupation). WEAT can be computed as:
%\vspace{-0.21cm}
\begin{align*}
\small
S(X, Y, A, B) &= \left[ \text{mean}_{x \in X} \operatorname{sim}(x, A, B) \right. \\
&\quad \left. - \text{mean}_{y \in Y} \operatorname{sim}(y, A, B) \right]
\end{align*}

\noindent{where} $X$ and $Y$ are sets of target pronouns $g$, and $A$ and $B$ are attribute bias words (\eg action verb). The idea is to measure the association strength between the target sets and the attributes.

\begin{table*}[t!]
\centering
%\small 
\resizebox{\textwidth}{!}{
\begin{tabular}{lcccccccccccccccc}

\hline 

& \multicolumn{8}{c}{WinoBias} & \multicolumn{8}{c}{WinoGender} \\ 
 \cmidrule(lr){2-9}\cmidrule(lr){10-17} 
& \multicolumn{4}{c}{verb} & \multicolumn{4}{c}{occupation} & \multicolumn{4}{c}{verb} & \multicolumn{4}{c}{occupation} \\ 
 \cmidrule(lr){2-5}\cmidrule(lr){6-9} \cmidrule(lr){10-13}\cmidrule(lr){14-17}
Model &  M & W & KL & WEAT &  M & W  & KL & WEAT & M & W  & KL &WEAT &  M & W & KL & WEAT \\ 

\hline 
  \addlinespace[1mm]
Human (all)  & 0.49  &  0.50    &  &   & &&& &0.58 & 0.41  \\
  \addlinespace[1mm]
BERT (all) & 0.78 & 0.21 &  &   & &&& &0.79 & 0.20 \\
ChatGPT (all) &0.70  &  0.29 &    & & && & &  0.88 & 0.11  &  \\  %115
\textcolor{black}{GPT-4} (all)  &0.69  &  0.30 &  &  &   & & &  &0.71 & 0.28 \\
\textcolor{black}{GPT-4-Trubo} (all)  &0.75  &  0.24 &  &  &   & & &  & 0.67    & 0.32   \\ %595/190
\textcolor{black}{GPT-4o} (all)  &0.66  &  0.34 &  &  &   & & &  & 0.71    & 0.28   \\ %595/190

\hline 
 \addlinespace[1mm]
GPT-2$_{\text{XL}}$   &  0.77 &  0.22  &  &   & &&& &0.76 & 0.23  \\

+ w2v &  \ccw{\NEW{0.88}}  &  \NEW{0.12} & \NEW{0.0088} & \NEW{0.088} &  \ccw{\NEW{0.61}} &  \NEW{0.39} &  \NEW{0.0087}&  \NEW{0.274} &  \ccw{\NEW{0.92}} & \NEW{0.08} & \NEW{0.0044} & \NEW{0.259} &  \ccf{\NEW{0.61}} & \NEW{0.39} & \NEW{0.0043} & \NEW{0.393}\\

+ GloVe &  \ccf{0.93}  &  0.07  & \NEW{0.0168} & \NEW{0.107} &  \ccff{\NEW{0.70}}  & \NEW{0.30} & \NEW{0.0099} & \NEW{0.326}  &  \ccf{\NEW{0.89}} & \NEW{0.11} & \NEW{0.0044} & \NEW{0.244} &  \ccw{\NEW{0.67}} & \NEW{0.33} & \NEW{0.0040} & \NEW{0.384} \\ % 'effect_size': 1.1015104549136863

+ fasttext  &   \ccf{\NEW{0.81}} & \NEW{0.19}   & \NEW{0.0053} &\NEW{0.012}  &   \ccw{\NEW{0.56}}&  \NEW{0.44} & \NEW{0.0066}& \NEW{0.111} &  \ccf{\NEW{0.89}} & \NEW{0.11} & \NEW{0.0013} & \NEW{0.027} & \NEW{0.49} &  \ccw{\NEW{0.51}} & \NEW{0.0019}  & \NEW{0.225}\\ % effect_size':  (occupation) 

+ GN-GloVe &   \ccw{\NEW{0.88}} & \NEW{0.12}  & \NEW{0.0141} &  \NEW{0.034} &     \ccw{\NEW{0.58}} & \NEW{0.42} & \NEW{0.0061}& \NEW{0.165}  &  \ccw{\NEW{0.92}} & \NEW{0.08} & \NEW{0.0053} & \NEW{0.273} & \NEW{0.39} &  \ccff{\NEW{0.61}} & \NEW{0.0031} & \NEW{0.331} \\  % 

+ DD-GloVe &   \ccw{0.92} & 0.08  & 0.0211  &  0.037  &  \ccf{0.54}  & 0.46  & 0.0042 &  0.153  &  \ccff{0.98} & 0.02 & 0.0070  & 0.244 & \ccff{0.68} &  0.32 &  0.0019 &  0.283 \\  % 

\hline 
\end{tabular}
}
\vspace{-0.2cm}
    \caption{Comparison of the bias amplification ratio results on benchmarks dataset between our proposed ClozeGender Score (CGS) against different baseline language models. Since the pronoun may occur in any part of the sentence, we take the mean probability of the full sentence with the specific pronoun. The results show that action verbs exhibit a higher amplified gender bias towards male entities compared to occupations. However, the occupation is more biased at the intrinsic word level with a higher  WEAT score.}

    \label{tb:benchmarkdataset}
\end{table*}

\noindent{\textbf{KL Bias Indication.} As we compare probabilities, we estimate the average change using KL divergence. A higher KL divergence indicates that the language model assigns significantly different probabilities to the two sentences, suggesting potential bias. Specifically, if the language model's similarity adjustment or bias update assigns different probabilities to sentences that are identical except for their gender pronouns, this indicates the presence of gender bias in the model.

\vspace{1.18pt}
\noindent{\textbf{Gender Bias Amplification Score.} We follow the work \cite{zhao2017men} that interprets gender bias towards a particular gender as a correction: %measure: 
\begin{equation*}
\small 
b(\text{co}, g) = \frac{c(\text{co}, g)}{\sum_{g' \in G} c(\text{co}, g')}
\end{equation*}

\noindent where \( c(\text{\textbf{co}ntext}, g) \) represents the occurrences of the context 
(\eg action verbs, object nouns, etc.) for each pronoun \(g \in G\), 
with \(G = \{m, w\}\).

\section{Experiments}

% fixed
To assess the proposed CGS score via GPT-2$_{\text{XL}}$ \cite{radford2019language}  we examine the most publicly available pre-trained word embeddings: word2vec \cite{Tomas:13}, GloVe \cite{pennington2014glove}, fasttext \cite{bojanowski2017enriching}, Gender Neutral models: (1) Static GN-GloVe \cite{zhao2018gender} and (2) Dictionary Definitions DD-GloVe \cite{an2022learning}. Our baseline are GPT-2$_{\text{XL}}$, ChatGPT \cite{OpenAI} GPT-4, GPT-4-Turbo \cite{achiam2023gpt} and GPT-4o \cite{openai2024gpt4o}  and  BERT \cite{Jacob:19}.

\vspace{1.18pt}

\noindent{\textbf{GenderLexicon:}}  We employ CGS score  on two datasets \ie occupation and gender neutral with last-cloze pronoun completion task. The result in Table \ref{tb:main_result} shows that object nouns have a more amplified bias toward female pronouns; meanwhile, action verbs are biased toward male pronouns.

\R{Additionally, we evaluate the bias between occupation and social groups  (\ie ethnicity). Specifically, we rely on gender-neutral pronoun sentences ([last cloze: them]), and then we update the probability with social group bias \eg ([$1-\text{sim}(\text{black}, \text{chef})$]. This makes the score more sensitive to social group biases, such as occupation-related ones. } Table \ref{tb:race-bias} shows that there is an occupational bias towards marginalized groups across all models except GN/DD-GloVe. Therefore, we adopt GN-GloVe architecture and trained Race-Neutral Glove (RN-GloVe) (using seed list of 13 ethnicities from both White \eg European and marginalized groups \eg African) on one billion words to optimize the race-neutral during training.  Also, since the out-of-the-box embedding vector white/black can be used to represent the race/color \cite{zhou2022sense}. We combined the two vectors (\eg black and african)  when computing the similarity score. However, for RN-Glove (\ie trained for scratch) we use the vector directly to measure the bias.  We also rely on Knowledge-Based sense-embedding models as baseline SenseEMbedding \cite{iacobacci2015sensembed}. Our RN-GloVe model reduces bias in marginalized groups, competitively to sense embedding baseline without sense knowledge. However, DD-GloVe, which leverages automated seed word expansion, proves to be a more effective debiasing measure.

\noindent{\textbf{WinoBias:}} \R{In the GenderLex dataset, we follow a fixed structure by having the pronoun last in the sentence. However, to evaluate our CGS Score on different datasets, where the pronoun may occur in any part of the sentence, we take the mean probability of the full sentence with the specific pronoun.} We extracted 1570 sentences (786 are unique) containing verbs and occupations that related directly to the pronoun (task 2). Additionally, to more accurately measure the direct bias related to gendered pronouns and avoid assuming false negatives across different gender groups, we focus only on the gendered occupation associated with the \textit{direct} pronoun, while making the first entity gender neutral, \eg \q{The \st{developer} \texttt{person}\footnote{Using \texttt{someone} results in a similar male bias ratio.} argued with the designer and slapped him in the face}. Table \ref{tb:benchmarkdataset} shows that our model interprets that the action verbs are more male-biased than occupations. However, occupation is more biased at the word level, without sentence context, with a higher WEAT score. 

%\noindent{\textbf{WinoGender:}} We follow the same procedures mentioned above and we manually extract 356 (178 are unique) samples containing verbs and occupations directly associated with pronouns. Table \ref{tb:benchmarkdataset} shows a similar result, the verbs are more male gender-biased than occupations. However, the occupation result is influenced by the introduction of a gender neutral \textit{someone} entity to the gendered occupation. Therefore, the lack of occupational stereotype association relevant to the pronoun results in an inaccurate estimation of the bias \ie by relying only on the similarity score \eg \q{the technician informed \textit{someone}} that [he/she could pay with cash].

\noindent{\textbf{WinoGender:}} We follow the same procedures mentioned above and we manually extract 356 (178 are unique) samples containing verbs and occupations directly associated with pronouns. Table \ref{tb:benchmarkdataset} shows a similar result, the verbs are more male gender-biased than occupations. However, the occupation result is influenced by the introduction of a gender neutral \textit{someone} entity to the gendered occupation. Therefore, the lack of occupational association relevant to the pronoun results in an inaccurate estimation of the bias \ie by relying only on the similarity score \eg \q{the technician informed \textit{someone}} that [he/she could pay with cash].

\begin{figure}[
t]
\small 
\centering 

\includegraphics[width=\columnwidth]{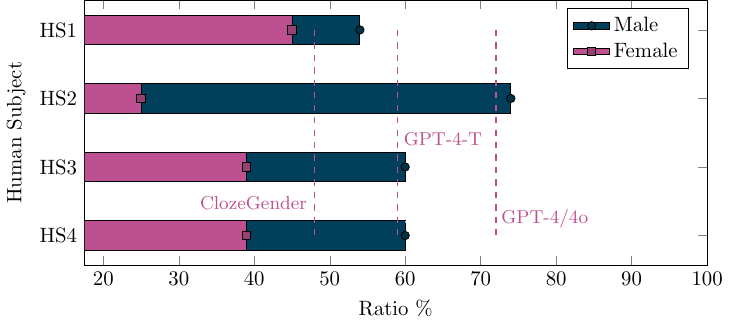}
\vspace{-0.6cm}

\caption{Human evaluation. We compare the professional annotations (Human Subjects (HS1)) with three random human subjects on the proposed GenderLexcion dataset. The average human subject (\ie HS2) has a similar bias to  GPT-4/4o, while our approach ClozeGender w/ GloVe has a lower bias compared to human bias.}

\label{fig:human-eva}
\end{figure}

\section{Human Evaluation}

\R{In this section, we compare our CGS score against human subjects and expert annotators. In particular,  we compare the professional annotation of Human Subjects (HS1) with three average human subjects on the proposed dataset.} We rely on two stages: the first evaluation stage is conducted without giving the human subject any instructions. The overall average bias is 80.74\% (bias towards male pronouns) in all occupations. The second stage is when we give the Human Subject the instruction to use the internet (\eg labor statistics) for unfamiliar occupations (\eg astronomer). Figure \ref{fig:human-eva}  illustrates the ratio after stage two, showing that with HS3 and HS4, the ratio drops 20\%. Note that increasing the number of human subjects results in the same 20\%  drop ratio after the second stage. \R{Therefore, humans introduce their own biases when occupations are unfamiliar to them \eg when the occupation is unclear, human subjects tend to apply a male bias. } The overall result demonstrates that HS3 and HS4 have aligned results to GPT-4-Turbo, whereas HS2 is similar to GPT-4/4o, while our approach (\ie ClozeGender Score (CGS) w/ GloVe) shows less bias when compared to HS1 (human bias). 

\begin{table}[!t]
\centering
\resizebox{\columnwidth}{!}{
\begin{tabular}{lcccc}
\hline & \multicolumn{4}{c}{CrowS-Pairs Stereotype} \\
\cline { 2 - 5 } Model & M & W & KL & WEAT \\
\hline 
 \addlinespace[1mm]

Human (all) & \new{0.53} & \new{0.47} & &  \\
   \addlinespace[1mm]

ChatGPT (all) & 0.54 & 0.45 \\ 
GPT-4 (all) & 0.57 & 0.42 \\

GPT-4-Turbo (all) &  0.47 & 0.52\\  % 119/129 2 run
 GPT-4o (all) &  0.48 & 0.51 \\  % 119/129 2 run
\hline 
 \addlinespace[1mm]
GPT-2$_{\text{XL}}$ \cite{radford2019language}  & \new{0.42} &  \ccw{\new{0.57}} &  \new{0.0068} &  \\

+ w2v &\new{0.49}   &  \ccf{\new{0.51}}  &  \new{0.0032} &  \new{0.245} \\
+ GloVe & \new{\ccw{0.52}} &  \new{0.48} & \new{0.0035} &  \new{0.229} \\ 
+ fasttext & \new{0.45} &  \new{\ccw{0.55}} & \new{0.0022} &  \new{0.087} \\ 
+ GN-GloVe & \new{0.42} &  \new{\ccff{0.58}} & \new{0.0029} &  \new{0.189} \\ 

+ \textcolor{black}{DD-GloVe} & \ccff{0.57} &  0.43 & 0.0021 &  0.108 \\ 

\hline
\end{tabular}
}
\vspace{-0.2cm}
\caption{{Result of our approach on CrowS-Pairs Stereotype dataset. Our CGS score leverages bias concepts extracted from stereotype pairs. GPT-4-T/GPT-4o has a similar bias to our model with w2v model with concept.}}
\label{tab:Stereotype}
\end{table}

\section{Gender Stereotype}
\R{In this section, we extend the evaluation of our ClozeGender Score (CGS) beyond gender bias to include gender stereotypes (\eg women/men can't drive).  We also demonstrate our approach on a non-English Japanese dataset.} Also, since the structure of stereotype datasets does not follow the entities corresponding to people referred to by their occupation (unlike previous datasets), measuring our contextual bias requires context. Therefore, we employ GPT-4 to extract a stereotype concept or topic from each sentence as the gender bias context.

\noindent{\textbf{CrowS-Pairs dataset \cite{nangia2020crows}}.} A crowdsourced dataset for measuring social biases, including gender and stereotypes. The dataset contains 1508 human-generated examples. We extract only the stereotype examples (498 samples) related to binary gender \eg  \q{[Men or Women] are bad at learning}, and then a stereo/anti-stereo concept is extracted   \eg \q{\textit{ineptitude}}. To reduce the human annotation time, the concept is generated using GPT-4 at first and then checked and corrected manually. Note that following the report on the inconsistencies in this dataset \cite{blodgett2021stereotyping}, we manually corrected and revised all the instances.

Table \ref{tab:Stereotype} shows that the initial bias via GPT-2 and most models tend to exhibit a more stereotypical female bias. The GN models further amplified the bias, as they only target de-biasing occupations. We evaluate the  ChatGPT/GPT 4 family with prompting by masking the gender, and without the concept, GPT-4-Turbo and GPT-4o have a similar bias ratio to our model with word2vec using the concept.

\begin{table}[t]
\centering
\resizebox{\columnwidth}{!}{
\begin{tabular}{lcccc}
\hline & \multicolumn{4}{c}{Japanese-Pairs Stereotype} \\
\cline { 2 - 5 } Model & M & F & KL & WEAT \\
 %\hline 

\hline 
 \addlinespace[1mm]
Human (all) & 0.50 & 0.50 & &  \\
 \addlinespace[1mm]

ChatGPT (all)  & 0.63 & 0.37 \\ % M198/W111 
GPT-4 (all)   & 0.38 & 0.61 \\ 
GPT-4-Turbo (all) & 0.23 & 0.76 \\  %w229/M71
GPT-4o  (all) & 0.25 & 0.65 \\
\hline 
\addlinespace[1mm]

GPT-Neo-JP-10B \cite{gpt-neox-library}& \new{0.37} &  \ccw{\new{0.63}} & \new{0.0003} &  \\
+ fasttext & \new{0.34}  & \ccff{\new{0.66}}  &  \new{0.0023} &  \new{0.015} \\
\hline

\addlinespace[1mm]

Mistral-JP-7B \cite{levine2024rakutenai} & \new{\ccff{0.81}} &  \new{0.19} & \new{0.0023} &  \\

+ fasttext & \new{\ccf{0.56}} &  \new{0.44} & \new{0.0034} & \new{0.015} \\
\hline

\end{tabular}
}
\vspace{-0.2cm}

\caption{Comparison of gender stereotype bias in Japanese across various LLMs. Our JP-7B CGS score better reflects human bias, while the  English-based Japanese model (10B) prioritizes English over Japanese, propagating bias similar to the GPT-4 family.}

    \label{tab:Stereotype-jp}
\end{table}

\noindent{\textbf{Pronouns Japanese-Pairs Stereotype:}} The pronouns in the Japanese language are gender-neutral and context-based rather than gender-specific. Therefore, the pronouns \textit{he} and \textit{she} can be used as context-dependent. For example, \textit{she/he} pronouns also mean boyfriend/girlfriend. To measure gender bias, we rely on the standard context-independent pronouns similar to \textit{he} and \textit{she} in English and male and female names. To create the dataset, we extract 15 stereotypical stereo/anti-stereo topics from traditional Japanese cultures for each gender as follows:
\noindent{\textbf{Man Stereotype:}} salaryman spirit, expressionless, breadwinner of the family, tech enthusiast, strong drinking capacity, shy around women, non-involved fathers, conservative, loyal to the company, lack of housework skills, sports enthusiast, modest, pressure to succeed, group-oriented, respect for elders.
\noindent{\textbf{Women Stereotype:}} housewife, submissive, fashion-conscious, office lady, dependent, good at cooking, modest, high-pitched voice, emphasis on youth, indirect communication, strong group awareness, cute culture, pressure to marry, family over career, elegant and polite.

\begin{table*}[t!]
\centering
%\small 
\resizebox{\textwidth}{!}{
\begin{tabular}{lcccccccccccccc}

\hline 

& \multicolumn{7}{c}{Secondary Bias Ratio } & \multicolumn{7}{c}{Main Bias Ratio} \\ 
 %\cmidrule(lr){2-9}\cmidrule(lr){8-13} 
  \cmidrule(lr){2-9}\cmidrule(lr){10-13} 
& \multicolumn{4}{c}{noun} & \multicolumn{4}{c}{verb} & \multicolumn{4}{c}{occupation} & \multicolumn{2}{c}{all/combined} \\ 
 \cmidrule(lr){2-5}\cmidrule(lr){6-9} \cmidrule(lr){10-13}\cmidrule(lr){14-15}
ClozeModel &  M & W & KL & WEAT &  M & W  & KL & WEAT & M & W  & KL & WEAT &  to-m & to-w  \\ 
\hline

\hline 
  \addlinespace[1mm]

Human &  &   &   &  & &  &  & & & & & & 0.54   &    0.45    \\
\hline
 
   \addlinespace[1mm]
    \addlinespace[1mm]
     \addlinespace[1mm]

GPT-J-6B \cite{gpt-j} &  &   &   &  & &  &  & & & & & & \new{0.53} & \new{0.46}\\

  \addlinespace[1mm]

+ fastext  & 0.41 & \ccw{0.59} & 0.0294 & 0.045 &   \ccw{0.53} & 0.47 & 0.0270  &  0.019 &  \ccf{0.51}  &  0.49 & 0.0323  & 0.070 &   0.48   &  0.52   \\ 
+ DD-GloVe   &0.49 & \ccf{0.51} & 0.0277 & 0.065 &   \ccf{0.51} & 0.49 & 0.0311  &  0.065 &  0.44  &  \ccw{0.56} & 0.0279  & 0.032 &   0.47   &  0.53      \\

 \addlinespace[1mm]
\hline 
  \addlinespace[1mm]
 \addlinespace[1mm]

LLAMA-3.1-8B \cite{llama3modelcard}     &  &   &   &  & &  &  & & & & & & \new{0.48} & \new{0.52}\\

  \addlinespace[1mm]

+ fastext  & 0.38 & \ccff{0.62} & 0.0368 & 0.055 &   0.47 & \ccw{0.53} & 0.0275 &  0.029 &  0.47  &  \ccf{0.53} & 0.0335  & 0.068 &   0.44   &  0.56   \\ 
+ DD-GloVe   & 0.44 & \ccw{0.56} & 0.0320 &   0.080 & 0.47 & \ccw{0.53}  &  0.0324 &  0.055  &  0.39 & \ccff{0.61}  & 0.0353 &   0.014   & 0.42 & 0.58  \\

 \addlinespace[1mm]
\hline 
  \addlinespace[1mm]
 \addlinespace[1mm]

DeepSeek-R1-8B \cite{deepseekai2025}     &  &   &   &  & &  &  & & & & & & 0.63 & \new{0.37} \\

  \addlinespace[1mm]

+ fastext  & \ccw{0.54} & 0.46 & 0.0688& 0.036 & \ccff{0.63} & 0.37 & 0.0727  &   0.011 &  \ccff{0.61}  &  0.39 & 0.0798  & 0.066 &   0.56   &  0.44   \\ 
+ DD-GloVe    & \ccff{0.60} & 0.40 & 0.0757 & 0.059 &  \ccff{0.63} & 0.37 & 0.0746  &  0.016 &  \ccw{0.55}  &  0.45 & 0.0682  & -0.015 &   0.59   &  0.41   \\ 
  \addlinespace[1mm]

\hline 
\addlinespace[1mm]
LLAMA-3.1-70B      &  &   &   &  & &  &  & & & & & & \new{0.51} & \new{0.49}\\

+ fasttext  & 0.38 & \ccff{0.62} & 0.0270 & 0.052 &   \ccf{0.51} & 0.49& 0.0204  &  0.021 &  \ccf{0.51}  &  0.49 &  0.0243  & 0.074 &   0.47   &  0.53   \\  
+ DD-GloVe    & 0.46 & \ccw{0.54} & 0.0242 & 0.064 &   0.48 & \ccf{0.52}  & 0.0261  &  0.010 &  0.41  &  \ccw{0.59} & 0.0259  & 0.001 &   0.45   &  0.55   \\ 

  \addlinespace[1mm]

\hline

\hline 
\end{tabular}
}
\vspace{-0.2cm}

\caption{\R{Evaluation results with ClozeGender Score via larger LLMs indicate that the RLHF-based model reduces bias in LLAMA-3.1-8B but amplifies it as the model size increases, similar to the behavior observed in the GPT-6B model without fine-tuning. The DeepSeek-R1 distillation from LLAMA-3.1-8B further reinforces these biases.}}

\label{tb:clozellms-all-full-paper}
\end{table*}

Since the GPT-4o tokenizer handles Japanese better than previous models, we employ this model to generate 10 sentences for each topic (in total 300 pairs), then we mask the pronoun and let the model predict the related stereotype gender. To apply our approach, we employ two Japanese LLMs models: (1) a model is fine-tuned on mixed dataset  GPT-NeoX-JP-10B \cite{gpt-neox-library}\footnote{\href{https://huggingface.co/matsuo-lab/weblab-10b}{https://huggingface.co/matsuo-lab/weblab-10b}} and (2) a state-of-the-art  Mistral-JP-7B based model \cite{levine2024rakutenai} that is fine-tuned on a clean Japanese dataset. For word embedding, we use Japanese-trained fasttext \cite{grave2018learning}.  Note that the topic used to create these pairs is the bias context. For the intrinsic WEAT score, we measure the score across two groups (\ie context) socially desirable  \eg success, respect, \etc and socially undesirable \eg submissive, expressionless.

\begin{table}[t!]
    \centering
    \resizebox{\columnwidth}{!}{
    \begin{tabular}{lcccccccc}
        \hline
        & \multicolumn{4}{c}{ClozeLast} & \multicolumn{4}{c}{ClozeAll} \\
        \cline{2-5} \cline{6-9}
        Model & M & W & KL   & HB\%&  M & W & KL & HB\% \\
        \hline
        \addlinespace[1mm]

\multicolumn{9}{l}{CPS-GPT-2$_{\text{XL}}$ \cite{felkner2023winoqueer}\quad \quad \quad \quad \quad   W: 0.00} \\

        \addlinespace[1mm]
        \addlinespace[1mm]

          GPT-2$_{\text{XL}}$     & 0.51 & 0.49 & 0.0116 & 0.71 & 0.82 & 0.18 &0.0001  & 0.63 \\
           \addlinespace[1mm]
        +  w2v        &       \new{0.49}  &  \new{\ccw{0.51}}   & \new{0.0150}  & \new{0.72} & \new{0.45} & \new{\ccw{0.55}} & \new{0.0015} & \new{0.64} \\
        +  GloVe      &     \new{\ccw{0.52}}    &  \new{0.48}   & \new{0.0180}  & \new{\textbf{0.73}}  &  \new{\ccw{0.54}} & \new{0.46} & \new{0.0017} & \new{0.70}  \\
       
        +  fasttext   &     \new{\ccf{0.51}}    &   \new{0.49}  & \new{0.0182}  & \new{\bf{0.73}}   & \new{\ccf{0.53}} & \new{0.47}  & \new{0.0011} & \new{0.63}  \\

        +  DD-GloVe   &     0.41    &   \ccff{0.59}  & 0.0181  & 0.72 & 0.10 & 0.90  &0.0024  & 0.45 \\         
         \hline
  \addlinespace[1mm]
        \addlinespace[1mm]

        \multicolumn{9}{l}{CPS-GPT-J-6B \quad \quad \quad \quad \quad  \quad \quad \quad \quad \quad \quad \quad \quad \quad  W: 0.01} \\ 
          \addlinespace[1mm]
        \addlinespace[1mm]
    
          GPT-J-6B      & 0.53 & 0.47 & 0.0228 & 0.71 & 0.58 & 0.42  & 4$\cdot10^{-5}$  & 0.73 \\
           \addlinespace[1mm]
        +  w2v         & \ccw{0.51} & 0.49 &0.0281& 0.72 & 0.39 & \ccw{0.61} & 0.0015  & 0.60 \\ 
        +  GloVe       & \ccff{0.54} & 0.46 & 0.0328  & \textbf{0.73} & \ccw{0.52}  & 0.48 & 0.0016  & 0.68 \\
       
        +  fasttext    & \ccw{0.51} & 0.49  & 0.0323 & \textbf{0.73} & 0.50 & 0.50 & 0.0011  & 0.63 \\

        + DD-GloVe     & 0.44 & \ccw{0.55} & 0.0279 & 0.71 & 0.08 & \ccff{0.92} & 0.0029  & 0.44 \\    
          \hline

    \end{tabular}
    } 
\vspace{-0.2cm}

       \caption{\R{Results on ClozeLast and ClozeAll on occupation title GenderLax dataset. The results show that ClozeLast reflects Human Bias (HB) accurately. }}
\label{ablation_study_genderlex}
\end{table}

\begin{table}[t!]
    \centering
    \resizebox{\columnwidth}{!}{
    \begin{tabular}{lcccccccc}
        \hline
        & \multicolumn{4}{c}{ClozeLast} & \multicolumn{4}{c}{ClozeAll} \\
        \cline{2-5} \cline{6-9}
        Model & M & W & KL   & HB\%&  M & W & KL & HB\% \\
        \hline
        \addlinespace[1mm]

        \addlinespace[1mm]
        \addlinespace[1mm]

          GPT-J-6B (GN)   &  0.41 & 0.59  & 0.0006  & 0.60  & 0.39  & 0.61 & 2$\cdot10^{-6}$  & 0.67  \\
           \addlinespace[1mm]
        +  w2v   (GN)   & 0.39  & \ccw{0.61} & 0.0016 & 0.63 & 0.38 & \ccw{0.62}  & 0.0013  & 0.58 \\
        +  GloVe  (GN)    & 0.45  & \ccf{0.55} & 0.0024 & 0.66 & \ccw{0.52} & 0.48  & 0.0014  & 0.68 \\
       
        +  fasttext (GN)  & 0.42  &  \ccw{0.58} & 0.0025 & 0.65 & 0.49 & \ccw{0.51}  & 0.0009  & 0.62 \\
         
        + DD-GloVe (GN)  & 0.22  & \ccff{0.78} & 0.0039 & 0.53 & 0.08 & \ccff{0.92}  & 0.0029  & 0.44 \\
         \hline 
          \addlinespace[1mm]
 
                    \addlinespace[1mm]
         GPT-J-6B (PN)     & 0.65 & 0.35 & 0.0047 &0.58  & 0.82  & 0.17  & 4$\cdot10^{-5}$ & 0.62 \\
         \addlinespace[1mm]
        +  w2v   (PN)        & \ccw{0.57} & 0.43 & 0.0064 & 0.64 & 0.41 & \ccw{0.59} & 0.0013  & 0.65 \\
        +  GloVe  (PN)       & \ccff{0.60} & 0.40 & 0.0082 & \textbf{0.67} & \ccw{0.53} & 0.47 & 0.0014  & \textbf{0.69} \\
        +  fasttext (PN)     & \ccff{0.60} & 0.40 & 0.0080 & 0.63 & \ccw{0.52} & 0.48 & 0.0009  & 0.64 \\
        +  DD-GloVe  (PN)    & 0.45 & \ccw{0.55} & 0.0065 & 0.63 & 0.10 & \ccff{0.90} & 0.0026  & 0.51 \\
        \hline

    \end{tabular}
    } 
\vspace{-0.2cm}
        \caption{\R{Result on ClozeLast and ClozeAll on occupation title on GenderLex dataset shows that \texttt{person} (PN) is more biased toward male than (GN) \texttt{someone} entity.}}

\label{ablation_study_genderlex_person}
    
\end{table}

Table \ref{tab:Stereotype-jp} shows that the fine-tuned model results on mixed datasets (\ie less Japanese dataset)  aligned with the stereotype of the GPT-4 family. In contrast, the high-quality dataset fine-tuned state-of-the-art Japanese model tends to produce an opposite stereotype bias towards male pronouns. However, using the stereotype context results in a more accurate reflection of human bias (on average 70\%). The 7B model estimates the male bias more accurately than 10B, \eg the model reflects a direct human bias like male-tech enthusiasm, (\q{[she/he] is using the latest smartphones and tablets and is skilled in presentations}). \R{Previous work addresses gender bias in Japanese using simple templates for occupations \cite{kaneko2022gender} and LLM-generated dialogues \cite{zhao2024gender}, lacking cultural context. To the best of our knowledge, this is the first dataset for analyzing Japanese stereotypes in the cultural context, with a metric for evaluating gender bias and stereotypes.} 

\section{Discussion}

The main finding of this study is that the weight of gender bias relies on the occupation as the main source of bias. However, when shifting the weight on nouns or verbs the gender bias changes. Similarly, when substituting the occupation with gender-neutral \texttt{person} or \texttt{someone}, the main bias ratio is often shifted to the opposite gender. Also, we have found that, in three datasets, the action verb is more biased than occupation toward male pronouns. In addition, there is a high occupational bias between social groups (\ie white/black) \eg fasttext has only \textit{dancer} as black occupation. Another observation is when we omit the occupation, our model is less correlated with the human ratio, which indicates that humans also rely on occupation as the main bias. In addition, by replacing the occupation with a gender-neutral term such as \texttt{person}, the male bias is reduced only in object nouns, and the action verbs remain associated with male entities, similar to the findings in \cite{bailey2022people_men}.

Although our evaluation is more focused on extrinsic bias metrics, we rely on two intrinsic bias metrics (WEAT and KL) to measure the gender bias, and our finding matches \cite{goldfarb2020intrinsic}  that intrinsic bias metrics and extrinsic do not correlate in measuring gender bias. In addition, unlike previous work that computes the WEAT score without considering the contextual context, we rely on the attribute sets (\eg action verb) to extract the bias gender targets set (\eg pronouns). This leads to overlap between genders for the same occupation or action verb in different contexts, reducing the impact of associations and lowering the overall WEAT score.

\section{Ablation Study}

\noindent{\textbf{Implications of Larger ClozeLLMs.} \R{We examine CGS with larger sizes of LLMs, including the GPT-J-6B \cite{gpt-j},  model with less strict data training, LLAMA-3.1-8B,  3.1-70B \cite{llama3modelcard}, and LLAMA-based deepseek-8B  \cite{deepseekai2025}. Table \ref{tb:clozellms-all-full-paper} demonstrates that as model size increases, bias persists even with RLHF fine-tuning with LLAMA-3.1-70B, similar to the bias observed in GPT-J-6B model. Furthermore, the DeepSeek-R1 distillation process from LLAMA-3.1-8B further increases these biases.}

\noindent{\textbf{ClozeLast and ClozeAll.} We examine the design choices of this work and how this score reflects human bias. For this, we compare the proposed ClozeLast with a ClozeAll-based model. The ClozeAll method clozes each word at a time and then uses all previous contexts to estimate the cloze probability, excluding the first word in the sentence. Then, the average score of all predicted cloze probabilities is used as the final bias score. Table \ref{ablation_study_genderlex} shows that ClozeLast reflects human bias more than ClozeAll across all models. As a baseline, we compare our score with the inspired Crows-Pairs Score (CPS) LLM-based score \cite{felkner2023winoqueer}, which struggles to capture bias as shown in Table \ref{ablation_study_genderlex}.

%Also, as a baseline, we compare our score with the inspired  Crows-Pairs Score (CPS) LLM-based score \cite{felkner2023winoqueer}, which struggle to capture the bias as shown in Table \ref{ablation_study_genderlex}.

\noindent{\textbf{Gender neutral: Person vs. Someone}
\R{\citet{bailey2022people_men} show that \texttt{person} entity are associated with male than female entities in word embeddings. Table \ref{ablation_study_genderlex_person} shows that \texttt{person} indeed is more biased for occupation (including verbs, similar to their finding), however, in the WinoBias benchmark dataset in Table \ref{tb:benchmarkdataset}, \texttt{someone} as male bias as \texttt{person}.}

\section{Conclusion}
%We propose a new dataset and framework that examines the gender bias problem in a broader aspect than relying only on occupation as a main source of bias. Our findings show that action verbs and object nouns also contribute to the overall gender bias. Also, our framework simplifies the exploration of other types of stereotypes and biases.

We propose a new dataset and framework that examines the gender bias problem in a broader aspect than relying only on occupation as a main source of bias, revealing that verbs and nouns also contribute to overall gender bias. The framework further facilitates exploration of other stereotypes and biases.

\section*{Limitations}
The cloze probability relies on a fixed position of the gender pronoun in the sentence to estimate the probability more accurately. We tackle this problem by taking the average probability of the sentence with the pronoun to adapt this score to benchmark datasets. Another limitation is that the proposed model utilizes pre-trained static word embedding to interpret the bias,  making the score dependent on the quality of these embeddings. However, this is not a limitation in our case, as the main idea of this work is inspired by likelihood revision, which requires external new information for revision. Therefore, we need to introduce an additional context bias source as a stand-alone, without depending on contextual information from the sentence. In addition, we addressed this by adding gender-neutral word embedding in our evaluation. Also, adapting this score to different datasets required human annotations for manual data extraction, such as identifying action verbs that are directly related to the pronoun, as ChatGPT and GPT-4 struggled in performing these tasks. However, this can be tackled by running the sample multiple times (\eg self-consistency sampling checking), reducing the human annotation time.

\section*{Ethics Statement}

We consider gender in a strictly binary manner (\ie man and woman), which oversimplifies the intricate and multifaceted nature of human identity. It would be more accurate to view gender as a spectrum, encompassing numerous variations beyond just two categories. Future work should acknowledge and address this complexity. %Furthermore, as models \ie language models, and word embedding are trained on general datasets, it is anticipated that they may also carry other biases such as those related to religion and culture.

\section*{Acknowledgment}
This work has received funding from the EU H2020 program under the SoBigData++ project (grant agreement No. 871042), by the CHIST-ERA grant No. CHIST-ERA-19-XAI-010, (ETAg grant No. SLTAT21096), and partially funded by  HAMISON project.

\bibliography{anthology,custom}
\bibliographystyle{acl_natbib}

\appendix

\end{document}